\documentclass[10pt,twocolumn,letterpaper]{article}

\usepackage{iccv}
\usepackage{times}
\usepackage{epsfig}
\usepackage{graphicx}
\usepackage{amsmath}
\usepackage{amssymb}
\usepackage{booktabs}
\usepackage{dsfont}
\usepackage{balance}
\usepackage{pifont}
\usepackage{multirow}
\usepackage{multicol}
\usepackage[linesnumbered,ruled,vlined]{algorithm2e}

\usepackage[pagebackref=true,breaklinks=true,letterpaper=true,colorlinks,bookmarks=false]{hyperref} %

\iccvfinalcopy %

\def\iccvPaperID{4141} %

\ificcvfinal\fi

\usepackage{color}
\usepackage{mathtools}

\newcommand{\Eq}[1]  {Eq.\ (\ref{eq:#1})}

\newcommand{\Fig}[1] {Figure \ref{fig:#1}}

\newcommand{\Tbl}[1]  {Table \ref{tbl:#1}}

\definecolor{purp}{rgb}{0.65, 0.16, 0.65}
\definecolor{greeen}{rgb}{0, 0.5, 0}

\DeclareMathOperator*{\argmin}{arg\,min}
\DeclareMathOperator{\Tr}{Tr}

\DeclarePairedDelimiter\floor{\lfloor}{\rfloor}

\begin{document}

\title{Deep Hough Voting for Robust Global Registration}

\author{Junha Lee \hspace{0.5cm} Seungwook Kim \hspace{0.5cm} Minsu Cho \hspace{0.5cm} Jaesik Park \vspace{1.5mm} \\
POSTECH CSE \& GSAI \vspace{1.5mm}\\
\small
\href{http://cvlab.postech.ac.kr/research/DHVR}{\url{http://cvlab.postech.ac.kr/research/DHVR}}
}
\maketitle
\ificcvfinal\thispagestyle{empty}\fi

\begin{abstract}
Point cloud registration is the task of estimating the rigid transformation that aligns a pair of point cloud fragments. We present an efficient and robust framework for pairwise registration of real-world 3D scans, leveraging Hough voting in the 6D transformation parameter space. First, deep geometric features are extracted from a point cloud pair to compute putative correspondences. We then construct a set of triplets of correspondences to cast votes on the 6D Hough space, representing the transformation parameters in sparse tensors. Next, a fully convolutional refinement module is applied 
to refine the noisy votes. Finally, we identify the consensus among the correspondences from the Hough space, which we use to predict our final transformation parameters. Our method outperforms state-of-the-art methods on 3DMatch and 3DLoMatch benchmarks while achieving comparable performance on KITTI odometry dataset. We further demonstrate the generalizability of our approach by setting a new state-of-the-art on ICL-NUIM dataset, where we integrate our module into a multi-way registration pipeline.
\end{abstract}

\section{Introduction}
\begin{figure}[t]
    \begin{center}
    \includegraphics[width=0.9\linewidth]{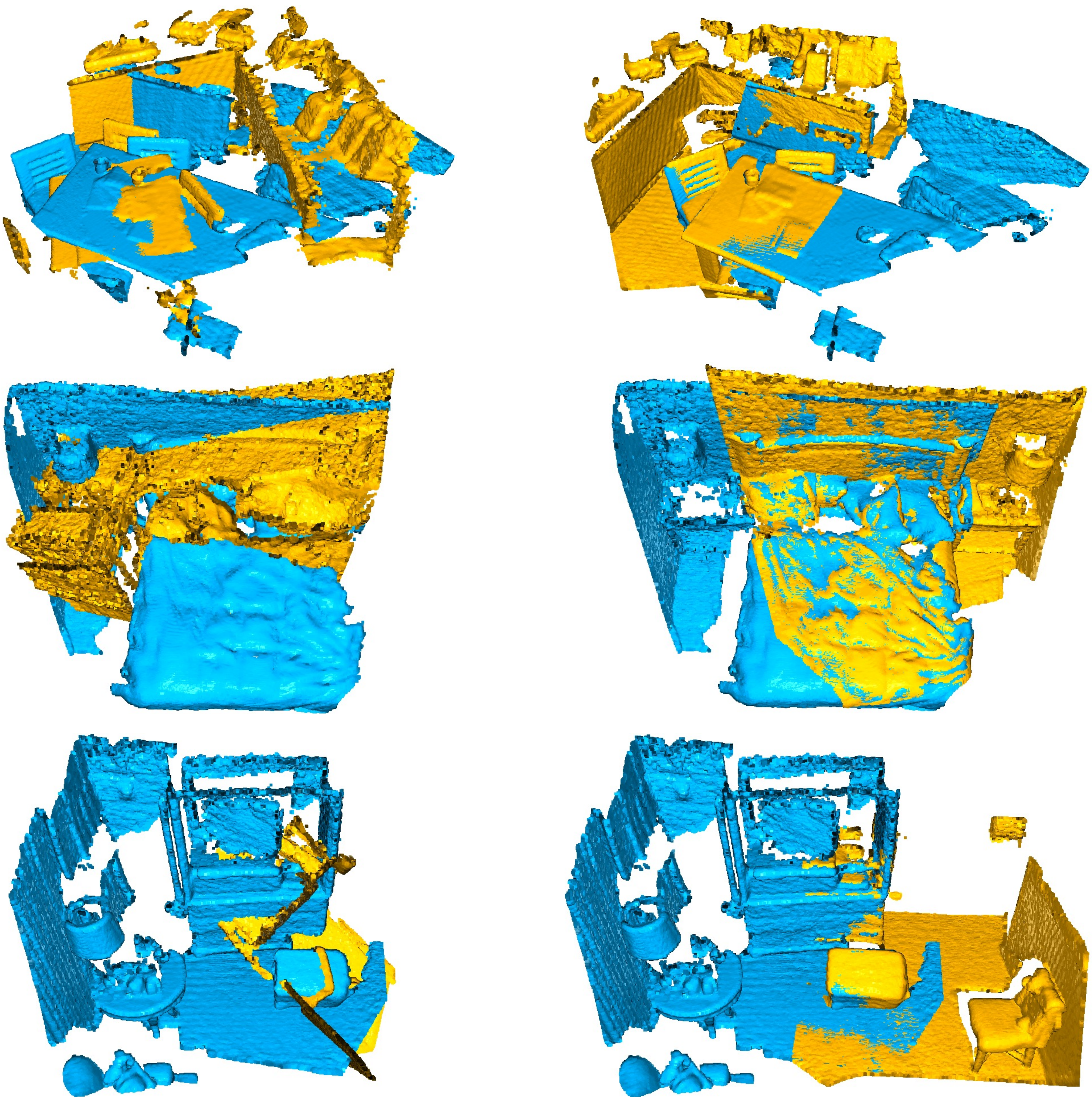}
    \end{center}
    \caption{Pairwise point cloud registration results of DGR~\cite{choy2020deep} (first column) and ours (second column) on 3DMatch dataset~\cite{zeng20163dmatch}. Our method is robust under low overlap and repetitive structures while being about twice as fast as DGR. 
    Best viewed in color.
    }
    \vspace{-0.3cm}
    \label{fig:teaser}
\end{figure}
Point cloud registration is one of the fundamental tasks in modern 3D computer vision, aiming to estimate the rigid transformation parameter to align a partially overlapping pair of point cloud fragments. 
It plays a vital role in autonomous driving, robotics, augmented reality, and other various applications.
The objective function of pairwise point cloud registration can be formulated as: 

\begin{equation}
\label{eq:objective_fn}
    E(\mathbf{R},\mathbf{t}) = \frac{1}{N} \sum_{(\mathbf{p},\mathbf{q}) \in \mathcal{C}} \| \mathbf{q} - \mathbf{R}\mathbf{p} -\mathbf{t} \|^2,
\end{equation}
where $\mathbf{p}, \mathbf{q} \in \mathbb{R}^3$ are points of a correspondence, $\mathcal{C}$ is the set of all true correspondences between the point cloud pair, N is the cardinality of $\mathcal{C}$, and $\mathbf{R} \in SO(3), \mathbf{t} \in \mathbb{R}^3$ are the rotation matrix and the translation vector, respectively. 
Therefore, point cloud registration is a dual problem of finding inlier correspondences and optimizing transformation parameters to minimize the geometric errors on these correspondences. 
Prior research on this topic can be categorized into geometric feature descriptors~\cite{bai2020d3feat,choy2019fully,deng2018ppf,deng2018ppfnet,gojcic2019perfect,rusu2009fast,tombari2010unique}, pose optimization algorithms~\cite{besl1992method,fischler1981random,zhou2016fast}, and the recent learning-based feature extraction and registration pipelines~\cite{aoki2019pointnetlk,choy2020deep,gojcic2020learning,wang2019deep,wang2019prnet}.
Despite the impressive performances of recent end-to-end pipelines, they either require a strong assumption on the high overlapping ratio of point clouds~\cite{wang2019deep}, lack scalability to be operable on large-scale point clouds~\cite{wang2019prnet}, or rely on iterative post-processing~\cite{choy2020deep}.

We propose a novel robust registration pipeline for large-scale point cloud fragment pairs to address these issues, yielding accurate rotation and translation predictions even in low point cloud overlap ratios.
The key idea is to view the registration problem as finding a consensus among the candidate correspondences on a \textit{discretized} 6D transformation space, leveraging Hough voting with sparse tensors. 
Using the noisy candidate correspondences, we accumulate votes into discrete bins of the Hough space. 
After all the votes are cast, we use a learnable refinement module on the Hough space to filter the noisy vote values.
Finally, the bin with maximum votes implies consensus among the candidate correspondences, which we use as our final prediction for the transformation parameters between the point cloud pair.
The Hough space is sparsely constructed to facilitate small-sized bins for higher accuracy while incurring significantly less memory overhead than a dense parameter space. 

The contributions of our work are as follows:
\vspace{-1.5mm}
\begin{itemize}
    \setlength\itemsep{-0.21em}
    \item We propose a novel approach for point cloud registration using Hough voting, where our \textit{sparse} 6D parameter space improves the accuracy and memory efficiency of our approach,
    
    \item We propose a learnable refinement module on the Hough space to enhance the representability of our sparse parameter space for increased robustness,
    
    \item We achieve a new state-of-the-art on the real-world indoor 3DMatch~\cite{zeng20163dmatch}/3DLoMatch~\cite{huang2021predator} datasets, and perform comparably on the outdoor KITTI~\cite{Geiger2012AreWR} dataset,

    \item We  prove our approach's generalizability by integrating our into a multi-way registration pipeline, evaluated on ~\cite{Handa2014ABF} to set a new state of the art,
    
    \item Our method exhibits faster speed compared to other learning-based registration pipelines.
\end{itemize}

\begin{figure}[t]
    \begin{center}
    \includegraphics[width=0.995\linewidth,page=1]{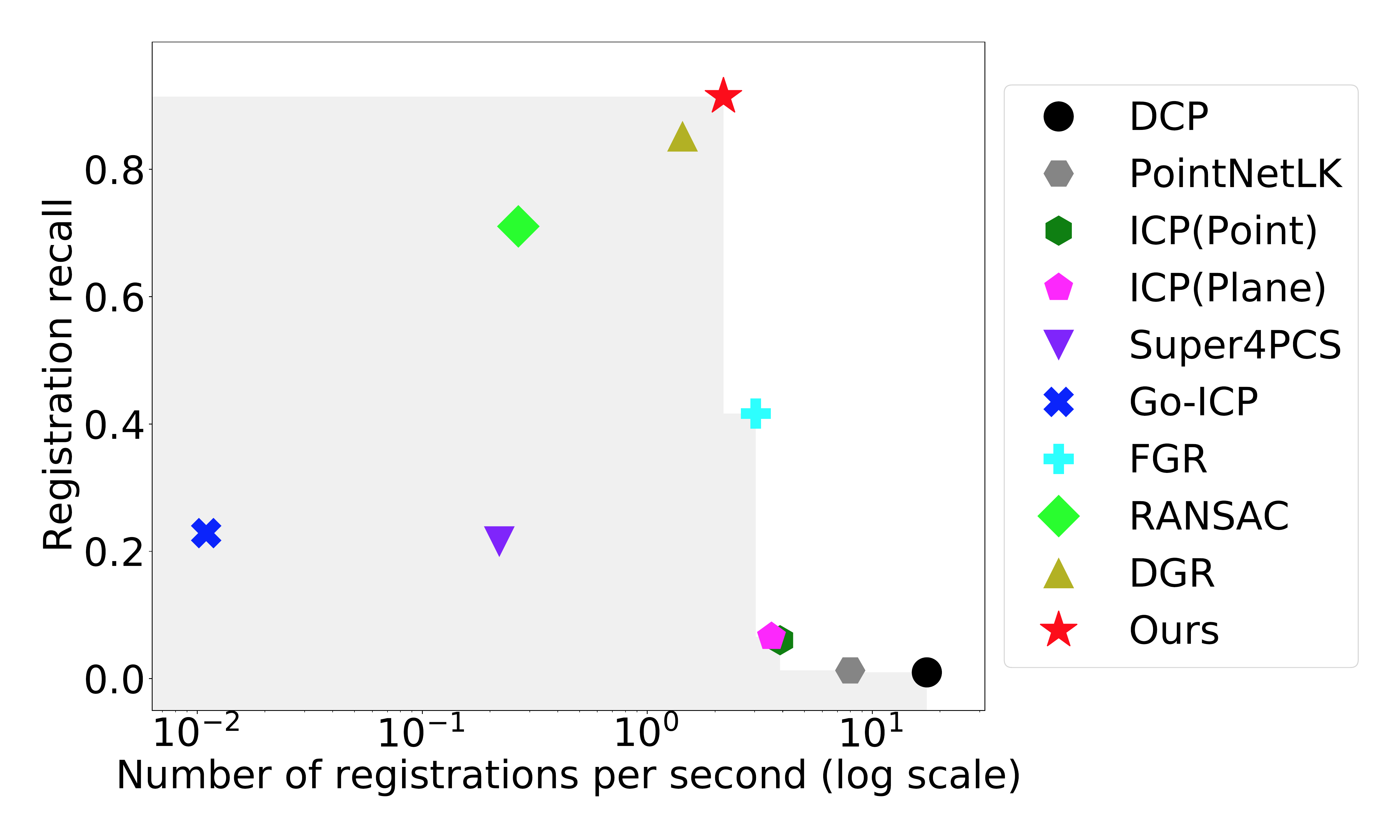}
    \end{center}
    \label{fig:pareto}
    \vspace{-0.3cm}
    \caption{Registration recall and the number of registration per second in log scale on  3DMatch benchmark. Our approach is the most accurate overall and fastest among learning-based methods.}
    \vspace{-0.3cm}
\end{figure}
\section{Related Work}
There are numerous approaches to solve point cloud registration. Prior work tackles the registration problem by decoupling the problem into a geometric feature description and robust matching problem.

\vspace{2mm}
\noindent
\textbf{Learning-based local feature.}
PointNet~\cite{qi2017pointnet} uses shared MLP network to encode individual coordinates of the input point cloud into high dimensional embedding space, and performs a permutation-invariant operation to aggregate features into a single global descriptor.
PerfectMatch~\cite{gojcic2019perfect} uses a smoothed density value (SDV) representation computed for each interest point, demonstrating rotation invariance and high generalizability. 
FCGF~\cite{choy2019fully} leverages a fully-convolutional 3D network and metric learning losses to extract features that outperform PerfectMatch while being up to 290 times faster.
However, these approaches focus mainly on feature descriptors, with little attention to feature detectors.
D3Feat~\cite{bai2020d3feat} therefore predicts detection scores with the features for each 3D point, showing that sampling features with higher scores is sufficient to achieve accurate and fast alignment.
These learning-based approaches outperform hand-crafted geometric features~\cite{spin,fpfh,pfh,shot,usc} which are sensitive to hyperparameter settings.

\vspace{2mm}
\noindent
\textbf{Robust model fitting.}
Correspondences obtained from feature matching often contain high proportions of outliers, which have to be filtered out for robust point cloud registration.
One of the most widely used methods is RANdom SAmple Consensus (RANSAC) and its variants~\cite{4pcongruent2008,pcl20153d,super4cps2014,fpfh,ransacpcd2007}, where a set of candidate correspondences are sampled every iteration, and an alignment is produced based on these correspondences to be evaluated.

Hough transform~\cite{hough1962method} was first introduced as a voting scheme to detect lines by discretizing parameter space into bins.
SIFT~\cite{sift2004} clusters features in pose space using Hough transform to estimate object poses. PHM~\cite{cho2015unsupervised} and HPF~\cite{min2019hyperpixel} applies this idea to efficient and robust matching algorithms for the task of object discovery and semantic correspondence, respectively. 
VoteNet~\cite{qi2019deep} votes for object centers in 3D space to predict 3D bounding boxes for object detection.
More recently, CHMNet~\cite{min2021chm} proposes to carry out local geometric matching using the idea of Hough voting to identify matches between images of the same category.

Our approach is inspired by recent methods using Hough voting, and we discretize the spatial transformation parameter space into sparse bins. 
We then vote on the sparse 6D space to predict the transformation parameters robustly based on the consensus from feature correspondences.

\vspace{2mm}
\noindent
\textbf{Global registration frameworks.}
Iterative Closest Points (ICP)~\cite{besl1992method} is a widely used approach due to its reliable performance in local registration. 
However, it requires an appropriate initial pose to avoid local minima and performs expensive nearest-neighbor queries in their inner loops. 
Other approaches make use of branch-and-bound~\cite{reg2016maron}, semi-definite programming~\cite{global2003maciel}, or maximal clique selection~\cite{yang2019polynomial}. 
These methods exhibit high accuracy but require time-consuming iterative sampling or heavier computation as the inlier ratio decreases.
FGR~\cite{zhou2016fast} and SparseICP~\cite{sparseicp2013} use robust objective functions to filter out outliers during optimization.
While FGR shows much shorter registration times albeit comparable performances, it is sensitive to the configurations of hand-crafted feature~\cite{fpfh}. 

End-to-end registration frameworks perform the task of feature learning and pose optimization jointly in a single forward step. 
PointNetLK~\cite{aoki2019pointnetlk} combines PointNet~\cite{qi2017pointnet} global features with an iterative pose optimization method into a single recurrent deep neural network. %
Deep Closest Points (DCP)~\cite{wang2019deep} addresses local optima issues and other difficulties in traditional ICP~\cite{besl1992method}, but it has a strong assumption that the correspondences are distributed over the entire point set.
While PRNet~\cite{wang2019prnet} integrates a keypoint selection step to DCP, making it applicable to aligning point cloud pairs with partial overlap, it is not feasible to be applied to large-scale point sets due to computational complexity.
Gojcic~\etal~\cite{gojcic2020learning} utilizes FCGF, incorporating a set filtering network~\cite{zhang2019learning} and extending to multi-way registration problem.
Deep Global Registration~\cite{choy2020deep} utilizes a 6D sparse convolutional network~\cite{choy2020high} to predict the inlier probabilities of feature correspondences, which are then used as weights to predict the rigid transformation parameters.
However, it requires an iterative post-processing step for refinement.
PointDSC~\cite{bai2021pointdsc} explicitly incorporates spatial consistency for pruning outlier correspondences prior to registration.
Predator~\cite{huang2021predator} proposes the overlap-attention block for early information exchange between the latent encodings of the point cloud pair, being effective even under very low overlap ratio.

Our proposed method integrates a learning-based feature extractor together with the highly robust sparse Hough voting module and a single-shot refinement module.
Despite being simpler, without any iterative processing steps or intensive outlier rejection methods, our proposed method shows improved efficacy over prior methods on real-world large-scale 3D scans.
\begin{figure*}[t]
    \begin{center}
    \includegraphics[width=0.95\linewidth]{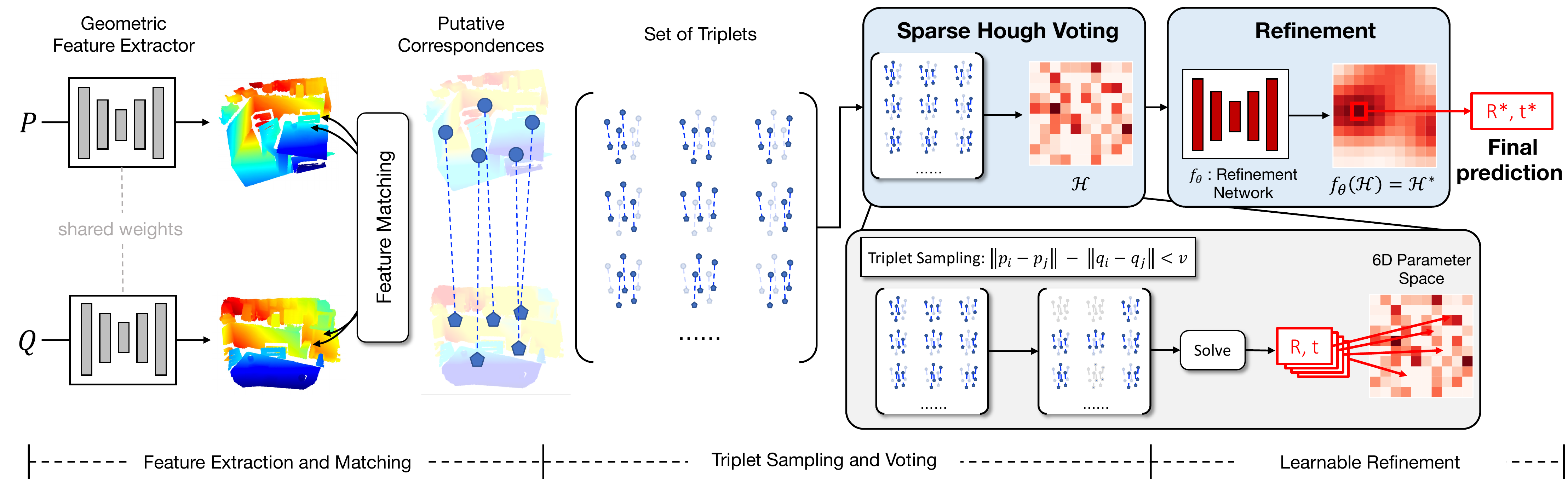}
    \end{center}
    \caption{Illustration of our proposed module. 
    We compute feature correspondences between point cloud pairs from locally extracted features, which are used to construct a set of triplets.
    Each triplet votes on the corresponding bin in the sparse Hough space in a parallelized fashion.
    We then perform a fully convolutional Hough space refinement to suppress the noisy Hough voting space.
    Finally, we identify the bin with the max votes, from where we make our final predictions on the transformation parameters.}
    \label{fig:overview}
\end{figure*}

\section{Method}
The proposed point cloud registration network is comprised of three steps: (1) local feature extraction, (2) sparse Hough voting, and (3) Hough space refinement.
Each step is detailed in the following subsections.

\subsection{Local Feature Extraction}

Given a point cloud pair, we extract local features that represent the geometric context of each point in metric feature space.
These features are used to compute putative correspondences between the point clouds using nearest neighbor search. 
For each point in a point cloud, the top-1 nearest neighbor is chosen as a candidate correspondence.
Therefore, given a point cloud pair with $\mathcal{M}$ and $\mathcal{N}$ points, respectively, we can obtain up to $\mathcal{M + N}$ unique candidate correspondences in this step.
Our work adopts Fully Convolutional Geometric Features (FCGF)~\cite{choy2019fully} as the local feature extractor.
Not only do features of FCGF show superior matching accuracy, but they also have low dimensionality of around 16 to 32, which facilitates the quick neighbor search.

Next, we construct a set of triplets for the subsequent Hough voting step, where each triplet consists of three unique candidate correspondences between the point cloud pair.
Motivated by FGR~\cite{zhou2016fast}, we perform simple tuple tests to filter out triplets with obviously spurious correspondences.
In our work, considering a triplet with correspondences $\mathbf{(p_1,q_1)}$, $\mathbf{(p_2,q_2)}$ and $\mathbf{(p_3,q_3)}$, the tuple test checks the following point-to-point distance conditions:
\begin{equation}
\label{eq:triangular_constraint}
    \forall i \neq j , {\displaystyle |\,} \|\mathbf{p}_i - \mathbf{p}_j\| - \|\mathbf{q}_i - \mathbf{q}_j\| {\displaystyle |\,} < 3 \times v ,
\end{equation}
where v is the spatial hashing size, or the voxel size.
The distance threshold $ 3 \times v$ is chosen as it was speculated to be the approximate upper bound to the error caused by spatial hashing.
This tuple test is effective under the invariance condition that rotation and translation transformations do not alter the point-to-point distances in a point cloud. 
While many other methods propose to filter out outliers more aggressively either by using iterative algorithms~\cite{chum2005matching,fischler1981random,fitzgibbon2003robust,hoseinnezhad2011m,raguram2012usac} or learning-based methods~\cite{brachmann2017dsac,choy2020high,moo2018learning,zhang2019learning}, we propose that our registration pipeline is highly robust to outliers, depreciating the necessity of additional, extensive outlier filtering. From here onwards, we use $\mathcal{C}$ to denote the set of triplets that passes \Eq{triangular_constraint}.

\subsection{Sparse Hough Voting}

We adapt Hough voting to discover the most sound consensus between the candidate correspondences to predict spatial transformation parameters. 
In our context, the spatial transformation parameters to predict are rotation and translation parameters in 3D space.
We use the axis-angle representation~\cite{goicp2015} to represent rotation in 3D space and three parameters along $x$-, $y$-, and $z$-axes for translation, having six degrees of freedom in total.
Therefore, we construct a discretized 6D parameter space, where each discrete unit is called a \textit{bin}.
Our objective is to identify the bin representing the transformation parameters that best represent the consensus among the candidate correspondences to be used as our final transformation parameter prediction.

Using axis-angle representation, we can represent a rotation as a 3D vector $\mathbf{r} \in \mathbb{R}^3$, where $\mathbf{r}/\|\mathbf{r}\|$ is the rotation axis and $\|\mathbf{r}\|$ is the rotation angle. We can convert a given rotation matrix $\mathbf{R}$ to its axis-angle representation $\mathbf{r}$ as,
\begin{equation}
\label{eq:rot_to_axisangle}
\begin{split}
    \|\mathbf{r}\| &= \arccos{(\frac{\Tr(\mathbf{R}) - 1}{2})}, \\
    \frac{\mathbf{r}}{\|\mathbf{r}\|} &= \frac{1}{2\sin{\theta}}\begin{bmatrix} \mathbf{R}_{32} - \mathbf{R}_{23} \\ \mathbf{R}_{13} - \mathbf{R}_{31} \\  \mathbf{R}_{21} - \mathbf{R}_{12}\end{bmatrix},
\end{split}
\end{equation}
where $\mathbf{R}_{ij}$ is the element in $\mathbf{R}$ in $i$-th row and $j$-th column, $\Tr(\cdot)$ is the matrix trace operator.
An axis-angle representation of 3D rotation is in the form of a sphere, while translation parameters can simply be represented in a cuboid space of $x$-, $y$-, and $z$-axes translations.
Together, the combined Hough space for rotation and translation would typically require a 6D space.
However, a dense and high-dimensional voting space is redundant, and translation parameters are unbounded in space, posing technical difficulties.

To this end, we use a \textit{sparse} 6D Hough space, leveraging the \emph{sparse tensor representation} provided by~\cite{choy20194d}.
Our implementation of sparse Hough space incurs minimal memory costs with no redundancy, as storage is required only for bins of the Hough space where votes are cast on.
This eschews the issue of translation parameters being unbounded as well.
Our sparse 6D Hough space $\mathcal{H}$ can represent all possible rotations and translations between two point clouds, where we can accumulate votes from the set of triplets onto the sparse bins. 
For each triplet, we need to retrieve the optimal rotation and translation parameters that minimize the geometric error illustrated in \Eq{objective_fn} as such:
\begin{equation}
\label{eq:objective_triplet}
    \mathbf{\hat{R}}_i, \mathbf{\hat{t}}_i = \argmin_{\mathbf{R},\mathbf{t}} \sum_{(\mathbf{p}, \mathbf{q}) \in \mathcal{C}_i}\|\mathbf{q} - \mathbf{R}\mathbf{p} - \mathbf{t} \|,
\end{equation}
where $\mathcal{C}_i$ is the $i$-th triplet.
We can retrieve $\mathbf{\hat{R}}_i$, $\mathbf{\hat{t}}_i$ by using Procrustes method~\cite{gower1975generalized}, a closed-form solution for rigid transformation in SE(3) as follows:

\begin{equation}
\label{eq:rot}
    \mathbf{\hat{R}}_i = \mathbf{V} \begin{bmatrix} 1 & 0 & 0 \\ 0 & 1 & 0 \\ 0 & 0 & \mathrm{det}(\mathbf{V}\mathbf{U}^T)\end{bmatrix} \mathbf{U}^T,
\end{equation}
where $\mathbf{\tilde{P}}_i\mathbf{\tilde{Q}}_i^T = \mathbf{U}\mathbf{\Sigma}\mathbf{V}^T$
is the singular value decomposition of a covariance matrix of $\mathbf{\tilde{P}}_i$, $\mathbf{\tilde{Q}}_i \in \mathbb{R}^{3 \times 3}$, zero-mean shifted coordinates that triplet $\mathcal{C}_i$ forms in each point cloud. The translation parameters are computed as:
\begin{equation}
\label{eq:tls}
    \mathbf{\hat{t}}_i = \mathbf{\bar{q}}_i - \mathbf{\hat{R}}_i \mathbf{\bar{p}}_i,
\end{equation}
where $\mathbf{\bar{p}}_i, \mathbf{\bar{q}}_i \in \mathbb{R}^3$ are the centroids of point cloud $\mathbf{P}_i, \mathbf{Q}_i$. We convert the obtained rotation matrix $\mathbf{\hat{R}}_{i}$ to its axis-angle representation $\mathbf{\hat{r}}_i$ as shown in \Eq{rot_to_axisangle}.

We then can use the obtained values of $\mathbf{\hat{r}}_i$ and $\mathbf{\hat{t}}_i$ to identify the bin, $\mathbf{v}_i \in \mathbb{N}^6$, in the Hough space for the triplet to vote on, simply by dividing the values of $\mathbf{\hat{r}}_i$ and $\mathbf{\hat{t}}_i$ by their corresponding Hough space bin sizes:
\begin{equation}
    \mathbf{v}_i = (\floor*{\frac{\mathbf{\hat{r}}_i}{b_{\mathbf{r}}}} \odot \floor*{\frac{\mathbf{\hat{t}}_i}{b_{\mathbf{t}}}}),
\end{equation}
where $b
_{\mathbf{r}}, b_{\mathbf{t}}$ are the hyperparameters denoting rotation and translation bin sizes, and $\odot$ is the vector concatenation operator.
In this work, we simply use 1's as the voting value for each triplet. 
Note that if we were to utilize the feature similarity values of the correspondences within each triplet for calculating the vote value, our proposed pipeline would be end-to-end trainable in theory.

Defining the bin sizes to be sufficiently small allows accurate predictions to be made, unlike large bin sizes where the range of rotations represented by a single bin is larger.
After all the votes have been cast, we end up with a 6D sparse Hough space of transformation parameters, with each bin holding its vote values from the voting process.
We refine this Hough space before making the final predictions on the transformation parameters.

\subsection{Learning to Refine Hough Space}

After all the votes have been cast, the sparse Hough space could be noisy due to triplets containing outlier correspondences.
Moreover, the quantization of continuous 6D transformation space into a discrete Hough space may introduce additional noise into the prediction pipeline without any refinement.
A simple aid to alleviate the noisy consensus would be to aggregate nearby votes by applying a simple Gaussian kernel.
A Gaussian kernel with an appropriate size may be sufficient to aggregate the local consensus and partially suppress the noisy votes. However, the receptive field would be limited to the kernel size, and it cannot exploit the high-dimensional geometric patterns inherent in the Hough space. 

To this end, we devise a high-dimensional sparse convolutional network $f_{\theta}: \mathbb{R} \to \mathbb{R}$, to refine the Hough space adaptively.
Specifically, $f_{\theta}$ is applied on the raw Hough space $\mathcal{H}$ and produces refined Hough space $\mathcal{H}^*$. 
We use a U-Shaped 6D sparse convolutional network as it can effectively increase the receptive field size for sparse tensors.
\begin{equation}
    \mathcal{H}^* = f_{\theta}(\mathcal{H}).
\end{equation}
To train $f_{\theta}$, we supervise it to perform a binary classification task where the true class is the bin which the ground truth transformation matrix falls into.
\begin{equation}
    \mathcal{L}_{bce}(\mathcal{H}^*,\mathbf{T}) = \sum_{i=0}^{|\mathcal{H}^{*}|}\left( y_i \log{\mathcal{H}_i^{*}} + (1-y_i)\log{\mathcal{H}_i^{*C}}\right),
\end{equation}
where $\mathbf{T}$ is the ground truth rigid transformation matrix, $y_i = 1$ if $i$-th bin of $\mathcal{H}^*$ is the bin that $\mathbf{T}$ falls into and otherwise 0 , $|\cdot|$ is the cardinality of a sparse tensor, and $\mathcal{H}_i^{*C} = 1 - \mathcal{H}_i^*$.
The overview of our proposed pipeline is illustrated in~\Fig{overview}.

\subsection{Comparison with RANSAC}
Here we elaborate on the differences between our method and RANSAC to highlight the design choice of the proposed algorithm.
While the motivation of determining the consensus among the candidate correspondences overlap between RANSAC and our method, there are distinct differences.
Unlike RANSAC, which adopts \textit{hypothesis-and-verify} pipeline and evaluates the hypothesis for every iteration, our method accumulates the consensus by utilizing the parallel voting procedure and evaluates the results only once after all voting has been finished.
This enables fast execution of the pipeline by removing the repetitive verification steps.
Furthermore, since our method accumulates the consensus in the form of a 6D sparse tensor, the fully-convolutional refinement network can be applied upon it, leveraging the inherent geometric bias in the accumulated consensus of 6D transformation parameter space.
We refer the readers to the Sec.\ref{section:experiment} for an empirical comparison between RANSAC and our method.
\section{Experiments}
\label{section:experiment}
We conduct experiments with our proposed approach in pairwise point cloud registration scenarios, and we further integrate our method into a multi-way registration pipeline.

\subsection{Datasets}
For indoor pairwise registration, we use 3DMatch benchmark~\cite{zeng20163dmatch}, consisting of point cloud pairs from various real-world indoor scenes.
Their ground truth transformations are estimated using RGB-D reconstruction pipelines~\cite{dai2017bundlefusion,Halber2017FinetoCoarseGR}.
While we use the provided train/test split, we follow the standard procedure~\cite{choy2019fully} to generate pairs with an overlap of as low as 30\% for both training and testing to evaluate the performance of our method under cases of low overlap ratio.
For pairwise registration of outdoor scenes, we use the odometry benchmark of KITTI LIDAR scans~\cite{Geiger2012AreWR}, consisting of real-world 3D scans taken using Velodyne laser scanners. 
We follow Choy \etal~\cite{choy2019fully} to create pairwise splits for training, validation, and testing.
To evaluate our method after integration into a multi-way registration pipeline, we use  ICL-NUIM dataset~\cite{Choi2015RobustRO} with simulated depth noise following ~\cite{choy2020deep}.

\subsection{Implementation}
Throughout our implementation, we extensively use CUDA to perform fast matrix multiplications and parallelized operations for higher efficiency.
In particular, the voting process of each correspondence set on all the bins of the Hough space is calculated in parallel.
For efficient nearest neighbor search when computing feature matches between point cloud pairs, we use Faiss~\cite{johnson2019billion}, an opensource library for efficient similarity search and clustering of dense vectors.
To utilize sparse tensor representation and perform sparse convolutions, we use MinkowskiEngine~\cite{choy20194d}, an auto-differentiation library for sparse tensors.
Open3D~\cite{Zhou2018Open3DAM} is widely used in our experiments for various classical registration methods and visualization of 3D data.

We use the official implementation of FCGF~\cite{choy2019fully} and the provided pre-trained weights for both 3DMatch and KITTI datasets. 
The Hough space refinement networks were trained for ten epochs with a batch size of 1, using an SGD optimizer with an initial learning rate of $10^{-3}$ and exponential learning rate decay factor 0.99.
We used spatial hashing~\cite{Rusu2009FastPF, Zhou2018Open3DAM} for 3DMatch and KITTI datasets with the size of 5cm and 30cm, respectively.
We use bin sizes of $b_\mathbf{r} = 0.02 \mathrm{rad}, b_\mathbf{t} = 0.02\mathrm{m}$ for 3DMatch dataset, and $b_\mathbf{r} = 0.005 \mathrm{rad}, b_\mathbf{t} = 0.1\mathrm{m}$ for KITTI in our experiments.

While leveraging the feature similarity values of correspondences could enable end-to-end training of our pipeline, we fix our feature extractor network in our experiments.
We brought the quantitative results of the baseline traditional and learning-based baseline methods from~\cite{choy2020deep}. 
For a fair comparison, we re-evaluate the~\cite{choy2020deep} on our environment to alleviate the performance difference due to the different hardware.
All the experiments are evaluated on an Intel Xeon Gold 5220R CPU, and an NVIDIA Titan RTX GPU, which is our best approximation of settings to ~\cite{choy2020deep}.

\subsection{Evaluation metric}
For all experiments, we report average Relative Rotation Error (RRE) and Relative Translation Error (RTE) as:
\begin{equation}
    \label{eq:rre}
   \mathrm{RRE}(\hat{\mathbf{R}}) = \arccos{\frac{\Tr(\hat{\mathbf{R}}^T\mathbf{R}^*)-1}{2}}
\end{equation}

\begin{equation}
    \label{eq:rte}
    \mathrm{RTE}(\hat{\mathbf{t}}) = \|\hat{\mathbf{t}}-\mathbf{t}^*\|^2_2,
\end{equation}
where $\hat{\mathbf{R}}$, $\hat{\mathbf{t}}$ are the predicted rotation matrix and translation vector, $\mathbf{R}^*$, $\mathbf{t}^*$ are the ground truth, and $\mathrm{Tr}(\cdot)$ indicates the trace of a matrix. 
Following ~\cite{choy2020deep}, we report \textit{recall}, which is the ratio of successful pairwise registrations. 
A pairwise registration is defined as successful if the RRE and RTE are below predefined thresholds. We compute the average RRE and RTE only on successful registrations for better numerical reliability.

However, average RRE and RTE are primarily affected by the recall and do not present an \textit{overall} view of how accurate the predictions are.
Therefore, we report the average RTE and RRE regardless of the success of the registration on 3DMatch dataset in Table \ref{tbl:rte_rre}.%

\subsection{Pairwise Registration}
We report pairwise registration results evaluated on the test set of 3DMatch benchmark~\cite{zeng20163dmatch} containing eight different scenes. 
We compare our methods with both classical methods and learning-based methods, including the current state-of-the-art.
In \Tbl{3dmatch}, we define the thresholds for RRE and RTE of successful registration to be 30cm and 15 degrees, respectively. 
Our method outperforms all the other techniques in terms of the recall, RRE, and RTE consistently in most scenes, achieving a new state-of-the-art.

\noindent
\textbf{Comparison with the traditional methods.}
As stated in ~\cite{choy2020deep}, methods of point-to-point and point-to-plane ICP, RANSAC~\cite{fischler1981random}, and FGR~\cite{zhou2016fast} implemented by Open3D~\cite{Zhou2018Open3DAM} were used.
For RANSAC and FGR, FPFH~\cite{fpfh} features were used.
Furthermore, Go-ICP~\cite{goicp2015} and Super4PCS~\cite{super4cps2014} were tested using open-source implementations in Python.
While the majority of the traditional methods fail in more cases, RANSAC methods demonstrate reasonable results, which improve with an increased number of iterations.
Our method is more than three times faster than RANSAC with 2M iterations and achieves significantly higher recall, RRE, and RTE.

\begin{figure}[t!]
    \begin{center}
    \includegraphics[width=0.995\linewidth]{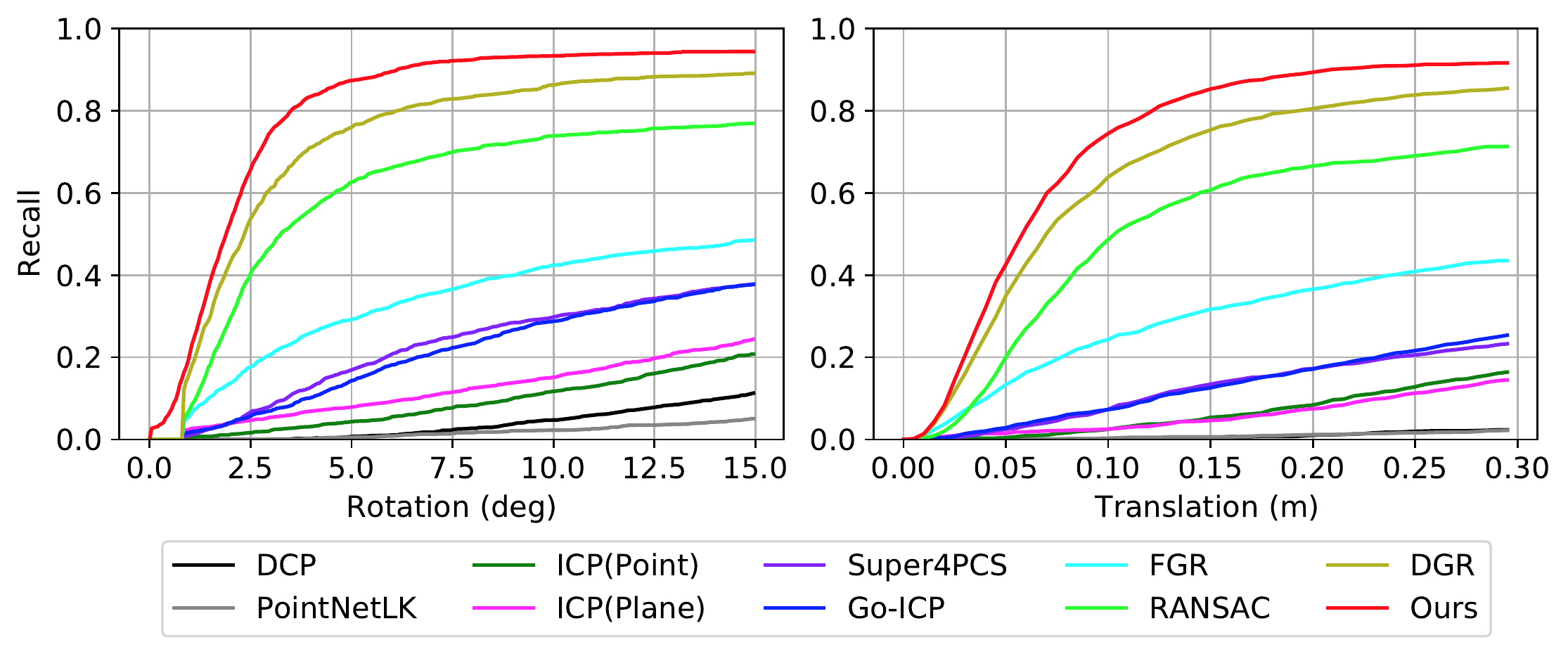}
    \end{center}
    \vspace{-0.3cm}
    \caption{Overall pairwise registration recall (y-axis) on 3DMatch benchmark with varying rotation (left image) and translation (right image) error thresholds (x-axis) for successful registration. Our approach outperforms baseline methods for all thresholds while being 2× faster than the most accurate baseline.}
    \label{fig:pr_3dmatch}
    \vspace{-0.3cm}
\end{figure}

\noindent
\textbf{Comparison with the learning-based methods.}
We compare our approach with learning-based pairwise registration methods, namely DCP~\cite{wang2019deep}, PointNetLK~\cite{aoki2019pointnetlk}, and DGR~\cite{choy2020deep}.
While other methods including 3DRegNet~\cite{dias20193dregnet} and PRNet~\cite{wang2019prnet} exist, they have shown to fail in ~\cite{choy2020deep}.
DGR offers to be the most effective among the compared methods. 
They previously achieved the highest recall and the lowest RRE and RTE while almost as fast as FGR. 
We show that our approach outperforms DGR in all evaluation metrics and shows to be around twice as fast as DGR as shown in \Tbl{3dmatch} and \Tbl{rte_rre}.
Furthermore, \Fig{pr_3dmatch} indicates that our method is robust under different rotation and translation thresholds.
\begin{table}[t!]
\small
\caption{Evaluation results on 3DMatch benchmark~\cite{zeng20163dmatch}. 
The first group of rows shows the results of classical global registration methods, and the second group of rows shows the results of ICP variants. The third group of rows shows the results of learning-based methods. 
Time includes feature extraction. The reported time of ~\cite{choy2020deep} on the original paper was 0.7s, whereas the recalibrated result was 0.96s.}
\vspace{0.2cm}
\label{tbl:3dmatch}
\centering
\resizebox{.45\textwidth}{!}{
  \begin{tabular}{l||c|c|c|c}
    \toprule
        & Recall (\%)    &  RTE (cm)  & RRE ($^\circ$)  & Time (s) \\ 
    \midrule
    FGR~\cite{zhou2016fast}             & 42.7    &  10.6   & 4.08    & 0.31    \\
    RANSAC-2M~\cite{ fischler1981random} & 66.1    &  8.85   & 3.00    & 1.39    \\
    RANSAC-4M                           & 70.7    &  9.16   & 2.95    & 2.32    \\
    RANSAC-8M                           & 74.9    &  8.96   & 2.92    & 4.55    \\ 
    \midrule
    Go-ICP~\cite{goicp2015} & 22.9    &  14.7   & 5.38    & 771.0   \\
    Super4PCS~\cite{super4cps2014} & 21.6    &  14.1   & 5.25    & 4.55    \\
    ICP(P2Point)~\cite{Zhou2018Open3DAM}            & 6.04    &  18.1   & 8.25    & 0.25    \\
    ICP(P2Plane)~\cite{Zhou2018Open3DAM}            & 6.59    &  15.2   & 6.61    & 0.27    \\ 
    \midrule
    DCP~\cite{wang2019deep}                     & 3.22    &  21.4   & 8.42    & 0.07  \\
    PointNetLK~\cite{aoki2019pointnetlk}        & 1.61    &  21.3   & 8.04    & 0.12   \\
    DGR~\cite{choy2020deep}                     & 85.2  &  7.73   & 2.58    & 0.70*    \\
    \midrule
    Ours                & \textbf{91.4} & \textbf{6.61} & \textbf{2.08} & 0.46 \\
    \bottomrule
  \end{tabular}
}
\end{table}

\begin{table}[t!]
\small
\vspace{2mm}
\caption{Mean and standard deviation of RTE and RRE on the entire 3DMatch benchmark~\cite{zeng20163dmatch}.}
\vspace{0.2cm}
\label{tbl:rte_rre}
\centering
\resizebox{.35\textwidth}{!}{
  \begin{tabular}{l||c|c}
    \toprule
        & RTE (cm) & RRE ($^\circ$) \\ 
    \midrule
    DCP~\cite{wang2019deep} & 1.41 $\pm$ 0.88  & 49.48 $\pm$  36.41 \\
    PointNetLK~\cite{aoki2019pointnetlk} & 2.28 $\pm$ 1.38  & 86.00 $\pm$ 47.70 \\
    ICP(Point)~\cite{Zhou2018Open3DAM} & 1.07 $\pm$ 0.87  & 35.06 $\pm$ 24.91 \\
    ICP(Plane)~\cite{Zhou2018Open3DAM} & 1.23 $\pm$ 4.48  & 35.32 $\pm$ 27.63 \\
    Super4PCS~\cite{super4cps2014} & 1.59 $\pm$ 1.52  & 64.64 $\pm$ 63.79 \\
    Go-ICP~\cite{goicp2015} & 1.28 $\pm$ 1.27  & 54.40 $\pm$ 58.19 \\
    FGR~\cite{zhou2016fast} & 1.08 $\pm$ 1.36  & 42.27 $\pm$ 52.31 \\
    RANSAC~\cite{fischler1981random} & 0.64 $\pm$ 1.18  & 26.25 $\pm$ 49.89 \\
    DGR~\cite{choy2020deep} & 0.30 $\pm$ 0.76  & 10.80 $\pm$ 29.40 \\
    \midrule
    Ours & \textbf{0.21} $\pm$ 0.67  & \textbf{7.95} $\pm$ 27.83 \\
    \bottomrule
  \end{tabular}
}
\end{table}
Feature extractors such as ~\cite{bai2020d3feat, choy2019fully}, integrated with RANSAC, are also applicable for the pairwise registration task on 3DMatch benchmark.
However, they use a different error threshold from ours and are not directly comparable to our method.
We include the comparison of our method with the feature extractors + RANSAC under their settings in the supplementary materials.

\begin{figure}[t!]
    \begin{center}
    \includegraphics[width=0.8\linewidth]{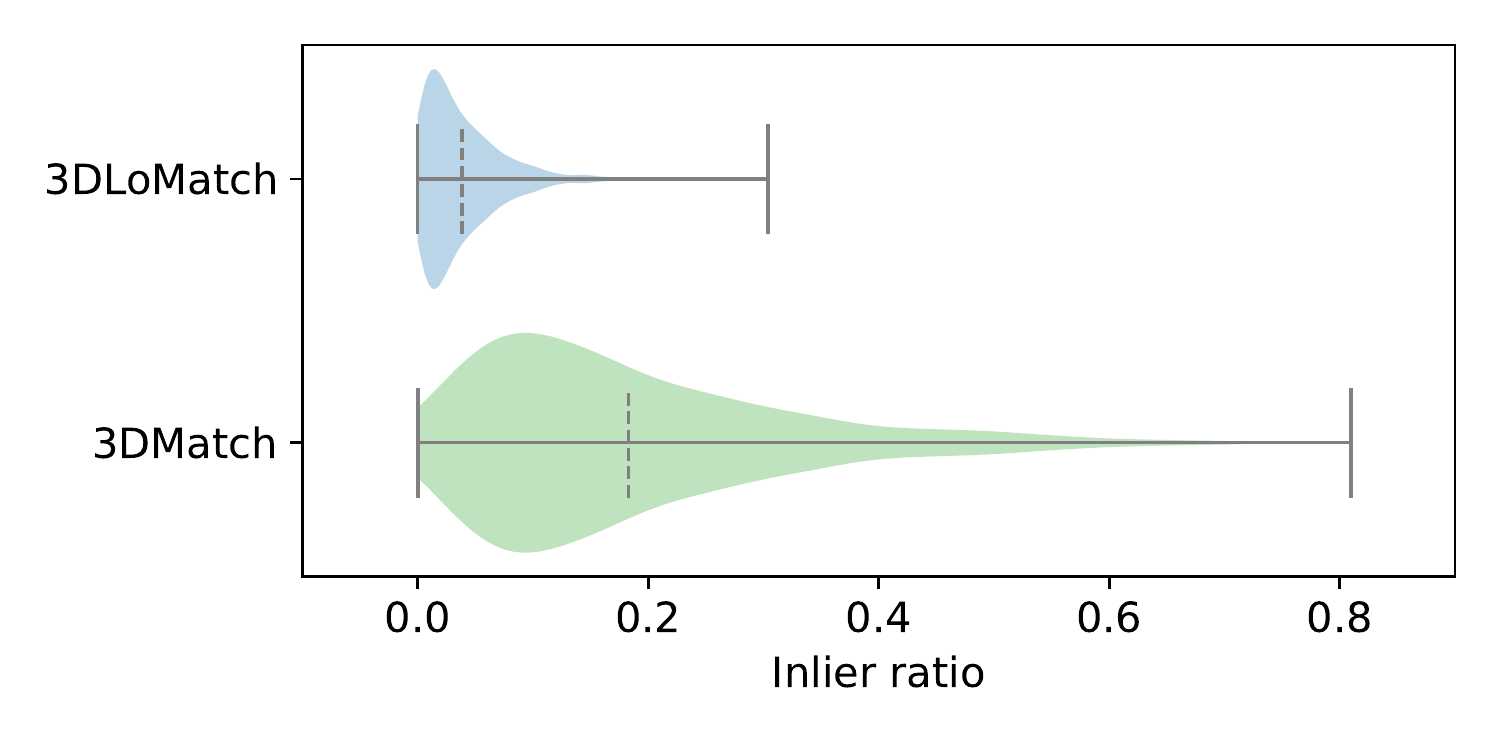}
    \end{center}
    \vspace{-0.3cm}
    \caption{The distribution of inlier ratio of the putative correspondences obtained by feature matching on 3DMatch~\cite{zeng20163dmatch} and 3DLoMatch~\cite{huang2021predator} datasets.}
    \label{fig:violin}
\end{figure}

\begin{figure}[t!]
    \begin{center}
    \includegraphics[width=0.9\linewidth]{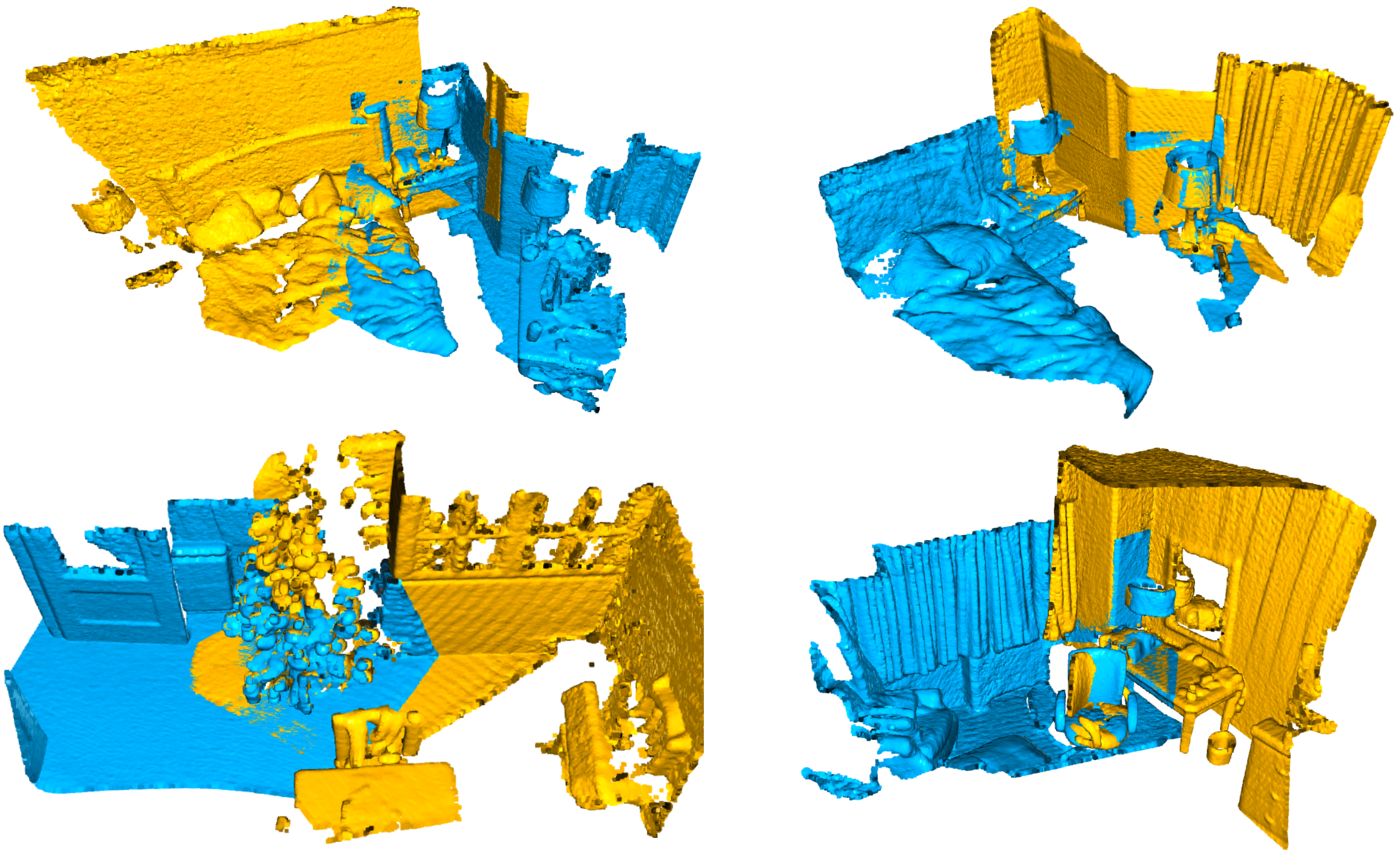}
    \end{center}
    \vspace{-0.3cm}
    \caption{Registration results on 3DLoMatch~\cite{huang2021predator} dataset. As can be also seen in \Fig{violin}, 3DLoMatch is more challenging than 3DMatch due to the lower overlap ratio. Best viewed in color.}
    \label{fig:3dlomatch}
\end{figure}

\noindent
\textbf{Robustness to lower inlier ratio.}
Furthermore, we show that our method is applicable for pairwise registration under even lower overlap ratios (10 to 30\%) using 3DLoMatch~\cite{huang2021predator} dataset.
\Fig{violin} shows that the inlier ratio of putative correspondences obtained by feature matching is significantly lower for 3DLoMatch dataset compared to 3DMatch dataset, therefore being a more challenging registration task.
The results are shown in \Tbl{3dlomatch}, where our method yields the best results using both FCGF~\cite{choy2019fully} and Predator~\cite{huang2021predator} features.
\Fig{3dlomatch} visualizes some of our registration results on 3DLoMatch dataset, where the low overlap ratio is clearly visible.

\begin{table}
\centering
  \caption{
  Registration recall on 3DLoMatch~\cite{huang2021predator} dataset. The results in the table excluding ours were taken from PointDSC~\cite{bai2021pointdsc}}
  \vspace{2mm}
  \label{tbl:3dlomatch}
  \resizebox{.32\textwidth}{!}{
  {\small
  \begin{tabular}{c|c|c}
    \toprule
    Feature & Method & Recall(\%)\\ 
    \midrule
    \multirow{3}{*}{FCGF~\cite{choy2019fully}} 
     & RANSAC                  & 35.7 \\
     & PointDSC~\cite{bai2021pointdsc} & 52.0 \\
     & Ours                    & \textbf{56.1} \\ \midrule
    \multirow{3}{*}{Predator~\cite{huang2021predator}} 
    & RANSAC                   & 54.2 \\
    & PointDSC~\cite{bai2021pointdsc} & 61.5 \\
    & Ours                     & \textbf{64.6} \\
    \bottomrule
  \end{tabular}
  }
  }
\end{table}

\subsection{Outdoor LiDAR Registration}
\begin{figure}[t]
    \begin{center}
    \includegraphics[width=0.995\linewidth]{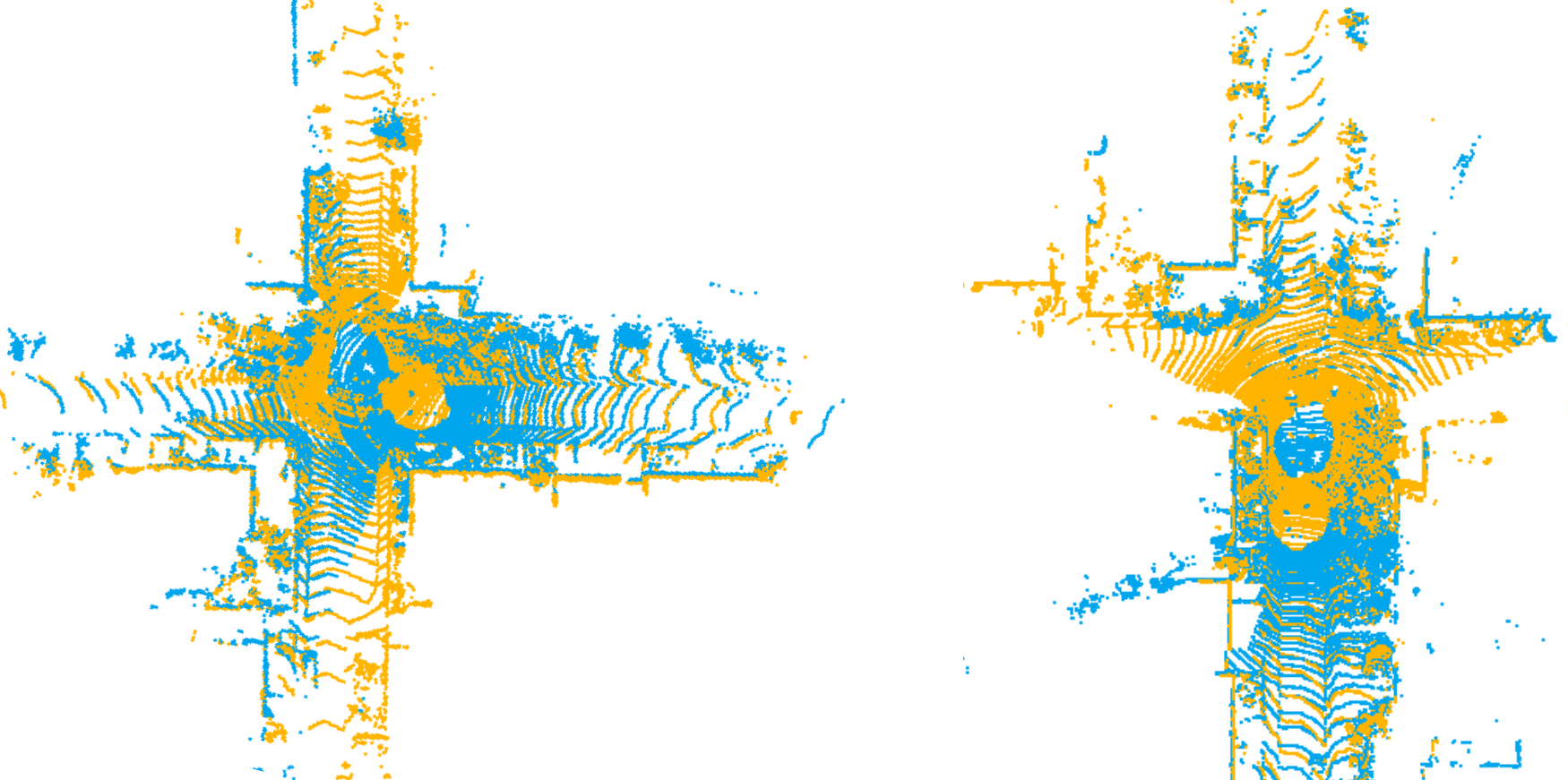}
    \end{center}
    \vspace{-0.3cm}
    \caption{Registration results of our method on  KITTI~\cite{Geiger2012AreWR} dataset. Best viewed in color.}
    \label{fig:kitti}
\end{figure}

We evaluate our method on outdoor LIDAR scans from KITTI dataset, following the evaluation protocols of ~\cite{choy2019fully}.
We use GPS-IMU to create the registration split of the dataset, containing pairs of point cloud fragments that are at least 10m apart.
The ground-truth transformations are generated using GPS, which is refined by ICP to fix errors in readings as in ~\cite{choy2019fully}. 
We use the FCGF network trained on the training set of the registration split to extract the pointwise features.
We use a voxel size of 30cm for downsampling the point cloud for this experiment since outdoor scenes tend to capture an enormous scope of data. 
The evaluation results are reported in \Tbl{kitti}, where our method shows to outperform the previous state-of-the-art in terms of \textit{recall} and RRE while being more than twice as fast as DGR.
FCGF + RANSAC achieves the lowest RTE, but our approach runs seven times faster in comparison.
The comparison of our method with the feature extractors~\cite{bai2020d3feat,choy2019fully} + RANSAC under different settings are in the supplementary materials for KITTI dataset as well. The visualization of the registration results is shown in \Fig{kitti}.
\begin{table}[t]
\centering
\small
  \caption{Evaluation results on KITTI test split~\cite{Geiger2012AreWR}. All methods are trained on 30cm voxel downsampled point clouds, and thresholds of 0.6m and $5^\circ$ are used to define successful registration. Time includes feature extraction.  The reported running time of ~\cite{choy2020deep} on KITTI is 2.29s, where the recalibrated result was 1.86s. 
  }
  \vspace{0.2cm}
  \label{tbl:kitti}
  \resizebox{.45\textwidth}{!}{
  \begin{tabular}{l||c|c|c|c}
    \toprule
                            & Recall (\%)    &  RTE (cm) & RRE ($^\circ$) & Time  \\ 
    \midrule
    \midrule
    FGR~\cite{zhou2016fast}          & 0.2     &  40.7   & 1.02    & 1.42    \\
    RANSAC~\cite{fischler1981random} & 34.2    &  25.9   & 1.39    & 1.37    \\
    FCGF+RANSAC~\cite{choy2019fully}        & \underline{98.2}    &  \textbf{10.2}   & \underline{0.33}    & 6.38    \\
    DGR~\cite{choy2020deep}          & 96.9   &  21.7   & 0.34    & 2.29*    \\
    \midrule
    Ours            & \textbf{99.1}     &  \underline{19.8}    & \textbf{0.29}     & \textbf{0.83}    \\ 
    \bottomrule
  \end{tabular}
  }
\end{table}
\vspace{-3mm}

\subsection{Time comparison}

While most of the quantitative results on 3DMatch and KITTI datasets are taken from~\cite{choy2020deep}, we re-evaluate~\cite{choy2020deep} on our environment for a fair comparison.
On 3DMatch dataset, the time reported on ~\cite{choy2020deep} is 0.70 seconds on average, while our re-evaluation on our environment yields 0.96 seconds. 
On outdoor LiDAR registration using KITTI dataset, the time reported on ~\cite{choy2020deep} is 2.29 seconds on average, while our re-evaluation on our environment yields 1.86 seconds on average. 
Therefore, the reported time in ~\cite{choy2020deep} is faster than our re-evaluated time on 3DMatch, but slower on KITTI. 
Our method consistently achieves a faster speed in comparison, demonstrating approximately two times shorter registration times on average.

\subsection{Ablation Study}

\Tbl{ablation_3dmatch} illustrates the results of ablation studies on our method regarding 1) the presence of Gaussian smoothing after voting, 2) the presence of the Hough space refinement module, and 3) the choice of voxel size for spatial hashing,
It can be seen that both Gaussian smoothing and refining consistently improve the performance for both voxel sizes of 5cm and 2.5cm.
However, it can be seen that these two schemes are not complementary, and using both Gaussian smoothing and Hough space refinement together does not constantly improve the results.
Applying the refinement module shows higher performance gains than applying Gaussian smoothing; therefore, we choose to use the learnable refinement module alone in our approach.

Using a smaller voxel size of 2.5cm demonstrates consistently superior results than using a larger voxel size of 5cm.
We conjecture that this is due to decreased errors derived from the spatial hashing step.
Nonetheless, our method sets a new state of the art even when using the larger voxel size of 5cm, proving the efficacy of our approach.

Due to spatial constraints, we refer the readers to the supplementary for additional ablation results regarding our method's execution time and memory requirements.

\begin{table}[t]
\centering
\small
  \caption{Ablation tests evaluated on 3DMatch benchmark~\cite{zeng20163dmatch}. 
  Ablations tests are carried out using the Gaussian kernel and the learnable refinement module for two different voxel sizes.}
  \vspace{0.2cm}
  \label{tbl:ablation_3dmatch}
  \resizebox{.45\textwidth}{!}{
  \begin{tabular}{c||c|c|c|c|c}
    \toprule
       Voxel Size  & Smoothing & Refine & Recall (\%)    &  RTE (cm) &  RRE ($^\circ$)  \\ 
    \midrule %
    \multirow{4}{*}{5cm}  &  &          & 86.7   &  7.59   & 2.45 \\
                          & \ding{51} & & 89.2   &  7.00   & 2.18 \\
                          & & \ding{51} & 91.4   &  \textbf{6.61}   & \textbf{2.08} \\
                          &\ding{51}  & \ding{51}   & \textbf{91.5}  & 6.89  & 2.23 \\
    \midrule
    \multirow{4}{*}{2.5cm}  &  &          & 90.5   &  6.69   & 2.03 \\
                          & \ding{51} &   & 91.7   &  6.25   & \textbf{1.88} \\
                          & & \ding{51}   & \textbf{93.7}   &  \textbf{6.23}   & 1.89 \\
                          &\ding{51}  & \ding{51}   & 93.1   & 6.23  & 1.94 \\
    \bottomrule
  \end{tabular}
  }
\end{table}

\subsection{Multiway Registration}
We show that our model can be integrated into a multi-way registration pipeline, exhibiting cross-dataset generalizability. 
Multi-way registration pipelines of RGB-D scans leverages pairwise registration to \textit{roughly} align 3D fragments.
This is followed by multi-way registration~\cite{Choi2015RobustRO} which optimizes the roughly aligned fragment poses with a robust pose graph optimization~\cite{g2ographoptimization}.

We replace the pairwise registration stage in a popular open-source implementation of a multi-way registration pipeline~\cite{Zhou2018Open3DAM} with our proposed method.
To illustrate the cross-dataset generalization abilities of our method, we use networks trained on 3DMatch training set while testing on the multi-way registration datasets.

We evaluate the results on the Augmented ICL-NUIM dataset~\cite{Choi2015RobustRO} with simulated depth noise for quantitative trajectory results, measuring absolute trajectory error (ATE) as the performance metric.
The results can be seen in Table \ref{tbl:multiview}, where we outperform state-of-the-art methods on all scenes except Living room 1, where we are second to DGR~\cite{choy2020deep}.
 
\begin{table}[t]
\centering
\small
  \caption{ATE(cm) on Augmented ICL-NUIM~\cite{Choi2015RobustRO}.}
  \vspace{0.2cm}
  \label{tbl:multiview}
  \resizebox{.45\textwidth}{!}{
  \begin{tabular}{l||c|c|c|c}
    \toprule
                            & FGR~[\color{green}51\color{black}]   & RANSAC~[\color{green}16\color{black}]  & DGR~[\color{green}8\color{black}] & Ours  \\ 
    \midrule %
    Living room1 & 28.01  & 23.32  & \textbf{21.06} & \underline{22.91} \\
    Living room2 & 31.71  & \underline{19.98}  & 21.88 & \textbf{16.37} \\
    Office1      & 13.89  & \underline{13.67}  & 15.76 & \textbf{12.58} \\
    Office2      & 15.61  & \underline{10.95}  & 11.56 & \textbf{10.90} \\
    \bottomrule
  \end{tabular}
  }
\end{table}

\section{Conclusion}

We present an efficient point cloud registration network for accurate and robust registration of large-scale, real-world 3D scans.
Our network pipeline is fast and straightforward, comprised of dense feature extraction and sparse Hough voting, followed by a learnable refinement module.
Features are extracted from a point cloud pair to compute putative correspondences. 
They are fed to our Hough voting module to vote on a transformation parameter space, constructed in a sparse manner to achieve high accuracy and low memory overhead.
We can find the consensus among the correspondences to predict the final transformation parameters.
Our sparse Hough voting module is shown to be highly robust against outliers, yielding state-of-the-art results on the indoor scene dataset and comparable results on the outdoor scene benchmark without integrating intensive outlier filtering methods.
We believe that integrating keypoint selection to reduce the number of initial correspondences or proposing triplet ranking for a smarter sampling of triplets would be an interesting future research direction for higher efficiency and performance.

\vspace{2mm}
\noindent
\textbf{Acknowledgement.} 
This work was supported by the NRF grant (NRF-2020R1C1C1015260, NRF-2017R1E1A1A01077999), the IITP grant (No.2019-0-01906, AI Graduate School Program - POSTECH funded by Ministry of Science and ICT, Korea), and SAMSUNG Electronics Co., Ltd.

{\small
\bibliographystyle{ieee_fullname}
\bibliography{egbib}
}

\clearpage
\renewcommand{\theequation}{a.\arabic{equation}}
\renewcommand{\thetable}{a.\arabic{table}}
\renewcommand{\thefigure}{a.\arabic{figure}}
\renewcommand*{\thefootnote}{\arabic{footnote}}
\renewcommand\thesection{\Alph{section}}
\setcounter{section}{0}
\setcounter{figure}{0}
\setcounter{table}{0}
\section{Appendix}
In this appendix, we provide additional details and results of our method.

\subsection{Refinement module architecture}
Here we describe the architectural details of the refinement networks used in the experiments.
We used the networks with the same architecture for all experiments.

\begin{figure}[ht]
\begin{center}
\includegraphics[width=0.9\linewidth]{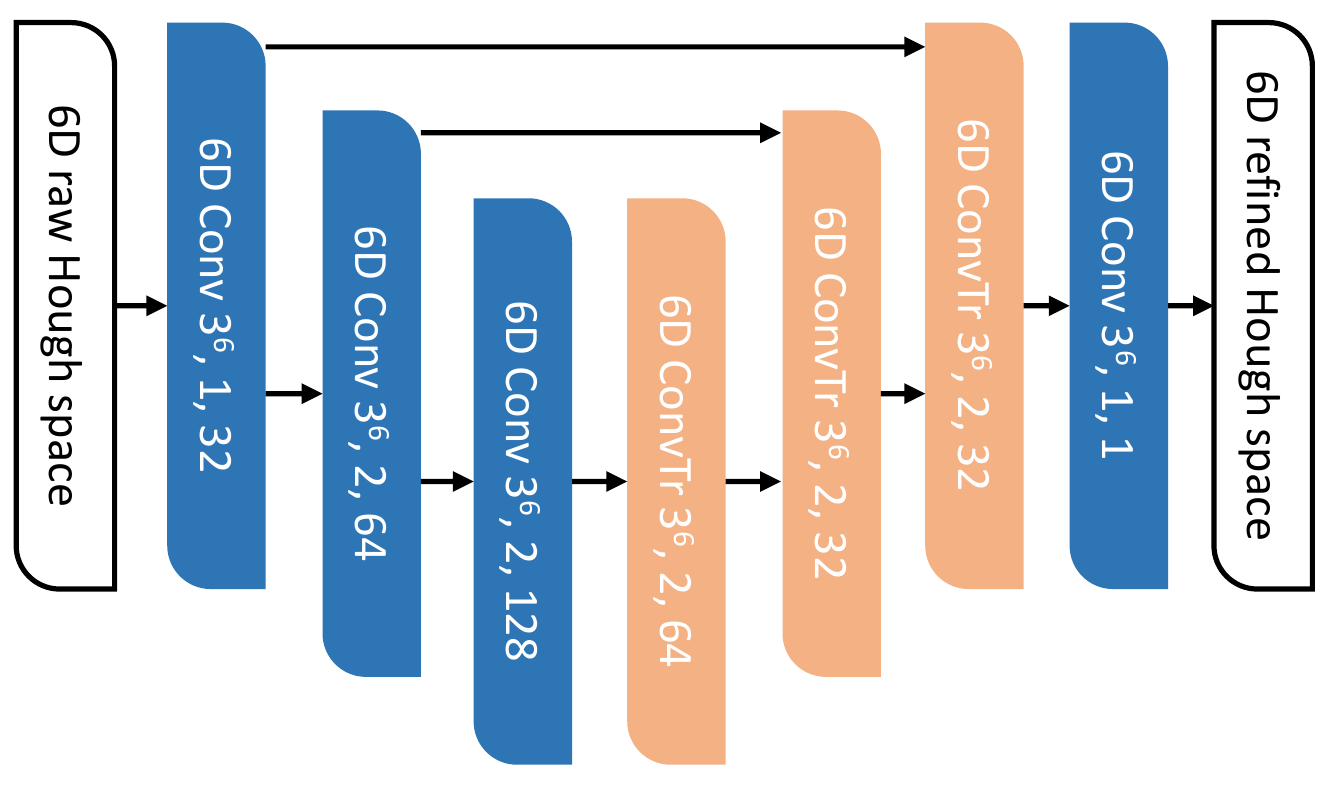}
\end{center}
\caption{Architecture of the refinement networks. The numbers in each box indicate the kernel size, stride, and the channel size. For example, the first convolutional layer has 6D kernel with kernel size 3, stride 1, and outputs 32 dimensional activations. }
\label{fig:architecture}
\end{figure}

As illustrated in \Fig{architecture}, the refinement networks are 6D U-shaped sparse convolutional networks with skip connections. 
They process the sparse, noisy 6D Hough space as the input to output the refined Hough space. 
They consist of 7 convolutional layers, with each layer followed by Batch normalization layer and ReLU activation except for the last convolutional layer. 
The strided convolutional/transposed convolutional layers respectively reduce/increase the cardinality of the input sparse tensor by the factor of 2. 
There are skip connections between the corresponding convolutional layers and transposed convolutional layers, concatenating the intermediate features along the channel dimension.

\subsection{Comparison with feature extractors}

In \Tbl{feature_extractors}, we compare our methods with the recent 3D feature descriptors and detectors under the different evaluation settings used in \cite{bai2020d3feat}.

Specifically, for 3DMatch~\cite{zeng20163dmatch}, the new condition for defining successful registration is $\mathrm{RMSE} < 0.2$m, whereas the condition that is used in our main manuscript was $\mathrm{RTE} < 0.3$m and $\mathrm{RRE} < 15^{\circ}$ following DGR\cite{choy2020deep}. 
For KITTI~\cite{Geiger2012AreWR}, the new thresholds for successful registration is $\mathrm{RTE} < 2$m and $\mathrm{RRE} < 5^{\circ}$, where the RTE threshold we used in our experiment was 0.6m, also following DGR\cite{choy2020deep}.
\begin{table}[h]
\centering
\small
  \caption{
    Comparison using the successful registration: $\mathrm{RMSE} < 0.2$m for 3DMatch, and $\mathrm{RTE} < 2.0$m \& $\mathrm{RRE} < 5^\circ$ for KITTI. Size indicates spatial hashing size used in cm. %
    }
  \vspace{0.1cm}
  \label{tbl:feature_extractors}
  \resizebox{.475\textwidth}{!}{
  \begin{tabular}{c|c|c|c|c}
    \toprule
    Dataset & Method & Size(cm) & Recall(\%) & Time(s)  \\ 
    \midrule
    \multirow{4}{*}{3DMatch~\cite{zeng20163dmatch}} 
    & PerfectMatch~\cite{gojcic2019perfect} & 1.875 & 80.3 & 23.0+      \\
    & FCGF~\cite{choy2019fully}        & 5 & \underline{87.3} &  1.70  \\
    & D3Feat~\cite{bai2020d3feat}       & 3 & 83.5 &  \underline{1.46}  \\
    & Ours & 5 & \textbf{92.9} &  \textbf{0.46}  \\
    \midrule
    \multirow{4}{*}{KITTI~\cite{Geiger2012AreWR}} 
    & 3DFeatNet~\cite{yew2018-3dfeatnet} & 20 & 96.0 & 15.0+     \\
    & FCGF~\cite{choy2019fully}       & 30 & 96.6 &  1.67 \\
    & D3Feat~\cite{bai2020d3feat}     & 30 & \textbf{99.8} &  \underline{1.45} \\
    & Ours & 30 & \underline{99.1} &  \textbf{0.83} \\
    \bottomrule
  \end{tabular}
  }
\end{table}

\subsection{Combination with FPFH feature}
In \Tbl{fpfh}, we compare our methods with baseline registration methods combined with FPFH feature descriptor. As results using FCGF descriptors are nearly saturated on 3DMatch, we evaluate our approach using FPFH~\cite{fpfh} without finetuning.
It can be seen that our method outperforms PointDSC~\cite{bai2021pointdsc} with comparable efficiency.
\begin{table}[h]
\small
\vspace{-0.1cm}
\caption{Evaluation on 3DMatch using FPFH feature descriptors. 
Time excludes feature extraction and matching.}
\label{tbl:fpfh}
\centering
\resizebox{.45\textwidth}{!}{
  \begin{tabular}{l||c|c|c|c}
    \toprule
        & Recall (\%)    &  RTE (cm)  & RRE ($^\circ$)  & Reg Time (s) \\ 
    \midrule
    FGR                                   & 40.67    &  9.83  &  3.99  & 0.28    \\
    RANSAC-10k                            &  60.63    & 11.79 &  4.35  & 0.55    \\
    RANSAC-100k                           & 77.20    & 7.42 &   2.62    &  5.25    \\
    3DRegNet                              &  26.31   & 9.60 &  3.75     &   0.05    \\
    DGR                                   & 69.13  &  10.80   &  3.78    &  2.49    \\
    PointDSC                              & \underline{78.50}   &  \textbf{6.57}  &  \underline{2.07}   &  0.09    \\
    \midrule
    Ours & \textbf{80.22} & \underline{6.87} & \textbf{2.06} & 0.10 \\
    \bottomrule
  \end{tabular}
}
\end{table}

\subsection{Additional scene-wise statistics}

\begin{figure*}
    \begin{center}
    \includegraphics[width=1.0\linewidth]{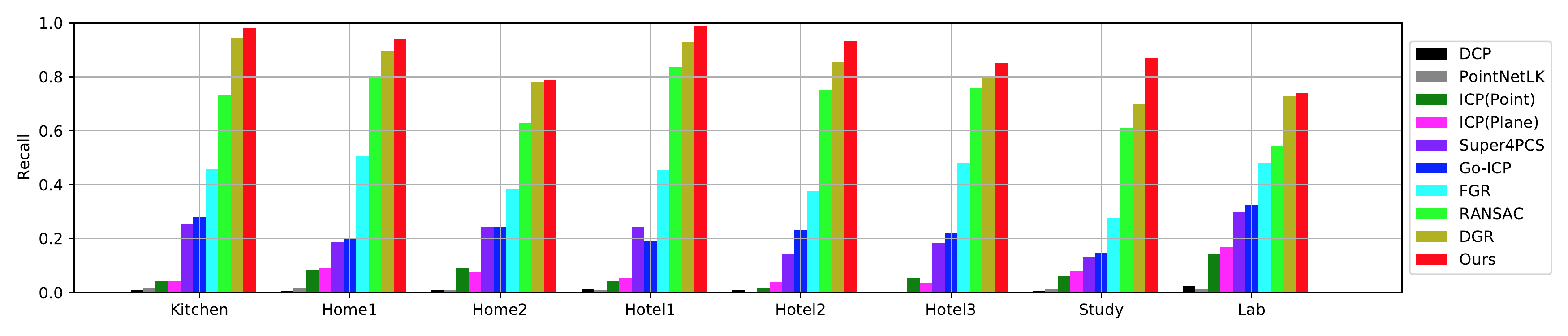}
    \includegraphics[width=1.0\linewidth]{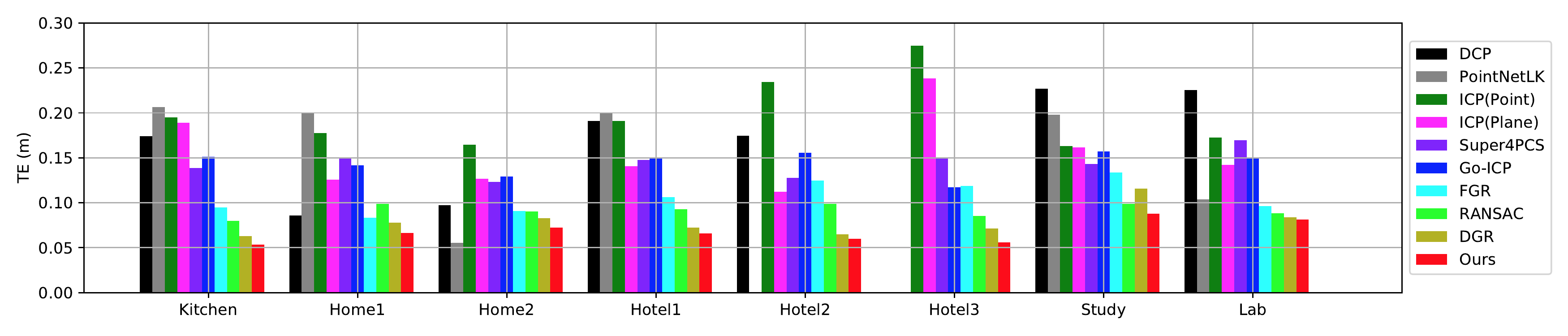}
    \includegraphics[width=1.0\linewidth]{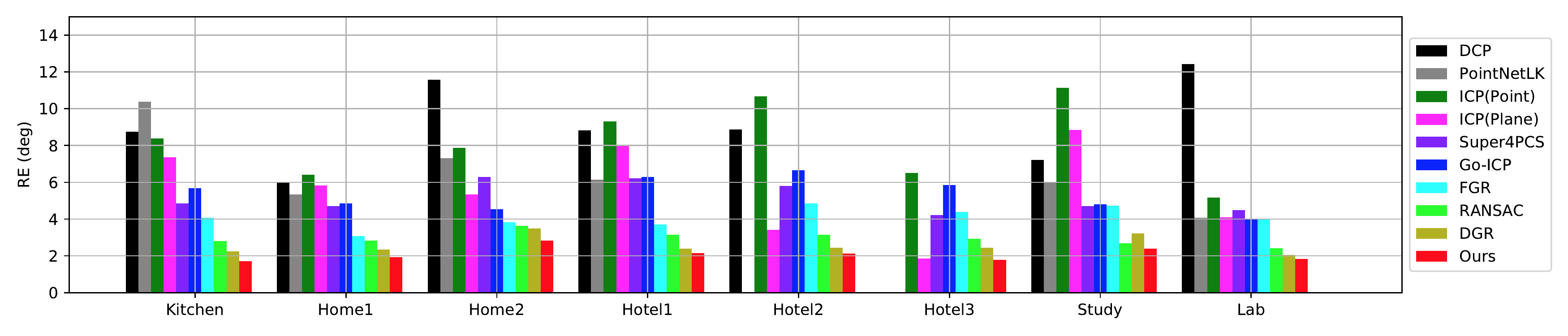}
    \end{center}
    \caption{Analysis of 3DMatch registration results per scene. The figure shows the recall, RE, and TE of each method in order. For recall, the higher the better; for RE and TE, the lower the better. Our approach is consistently better on all unseen test scenes. Note that a missing bar corresponds to zero successful alignments.}
    \label{fig:bar_3dmatch}
\end{figure*}

In \Tbl{3dmatch_scenewise}, the comprehensive scene-wise statistics for our best-performing model on the 3DMatch benchmark are presented.
We present a visual comparison of scene-wise statistics with baseline methods in \Fig{bar_3dmatch}. 
Our method consistently outperforms all baselines on all scenes.
\begin{table}[h]
\centering
\small
  \caption{Scene-wise statistics of our method on the 3DMatch~\cite{zeng20163dmatch} test set. The average is calculated for each pair of point clouds.}
  \vspace{0.2cm}
  \label{tbl:3dmatch_scenewise}
  \resizebox{.45\textwidth}{!}{
  \begin{tabular}{l||c|c|c|c}
    \toprule
                            & Recall(\%)  &  RTE(cm)  & RRE(deg)  & Time(s) \\ 
    \midrule
    Kitchen             & 98.0    &  5.35   & 1.72    & 0.42    \\
    Home1               & 94.2    &  6.63   & 1.93    & 0.47    \\
    Home2               & 78.9    &  7.23   & 2.84    & 0.51    \\
    Hotel1              & 98.7    &  6.58   & 2.14    & 0.48    \\ 
    Hotel2              & 93.3    &  6.01   & 2.12    & 0.44   \\
    Hotel3              & 85.2    &  5.60   & 1.78    & 0.46    \\
    Study               & 87.0    &  8.77   & 2.39    & 0.43    \\
    Lab                 & 74.0    &  8.15   & 1.83    & 0.69    \\ 
    \midrule
    Avg                 & 91.4    &  6.61   & 2.08    & 0.46    \\
    \bottomrule
  \end{tabular}
  }
\end{table}

\subsection{Analysis on memory usage}

In \Tbl{memory}, the analysis of memory usage of DGR~\cite{choy2020deep} and ours during evaluation is presented. Note that our learnable refinement module introduces slightly more memory usage but it leads to significant boost in the registration accuracy.
\begin{table}[h]
\centering
\small
  \caption{Analysis on peak memory usage and running time on 3DMatch benchmark~\cite{zeng20163dmatch}.}
  \vspace{0.2cm}
  \label{tbl:memory}
  \resizebox{.45\textwidth}{!}{
  \begin{tabular}{l||c|c}
    \toprule
       & Peak Mem. Usage (GB) & Time (s)  \\ 
    \midrule
    DGR~\cite{choy2020deep} & 3.500 & 0.96\\
    \midrule
    Ours                    & 2.949 & 0.33\\
    Ours + Gaussian smoothing           & 3.172 & 0.38\\
    Ours + Learnable refinement         & 4.344 & 0.46\\
    \bottomrule
  \end{tabular}
  }
\end{table}

\subsection{Additional qualitative results}

\begin{figure*}[t]
\begin{center}
\includegraphics[width=0.95\linewidth]{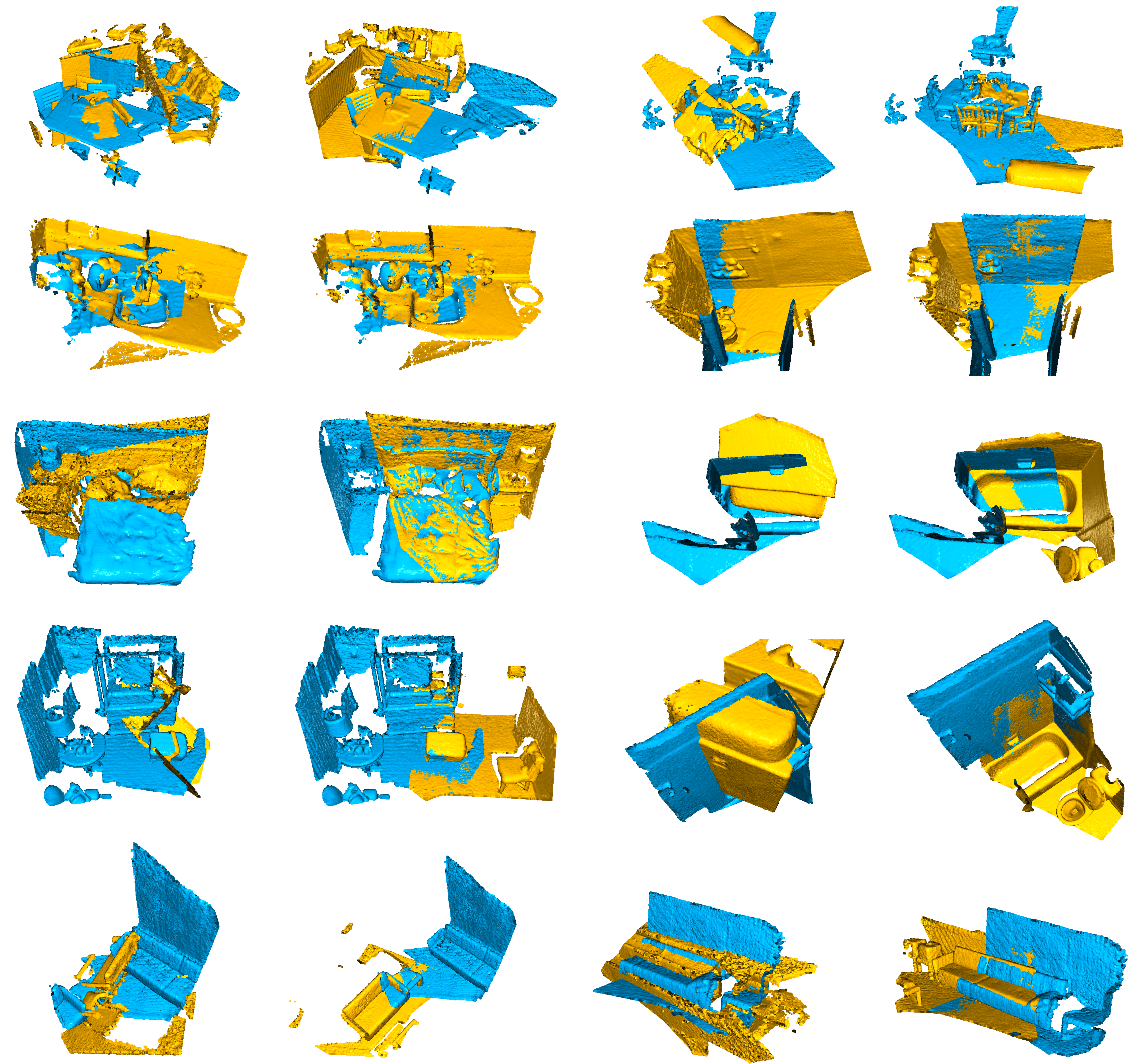}
\end{center}
\caption{Qualitative comparison of DGR~\cite{choy2020deep}(1st, 3rd column) and ours (2nd, 4th column) on 3DMatch benchmark.}
\label{fig:qualitative_3dmatch}
\end{figure*}

\begin{figure*}[t]
\begin{center}
\includegraphics[width=0.95\linewidth]{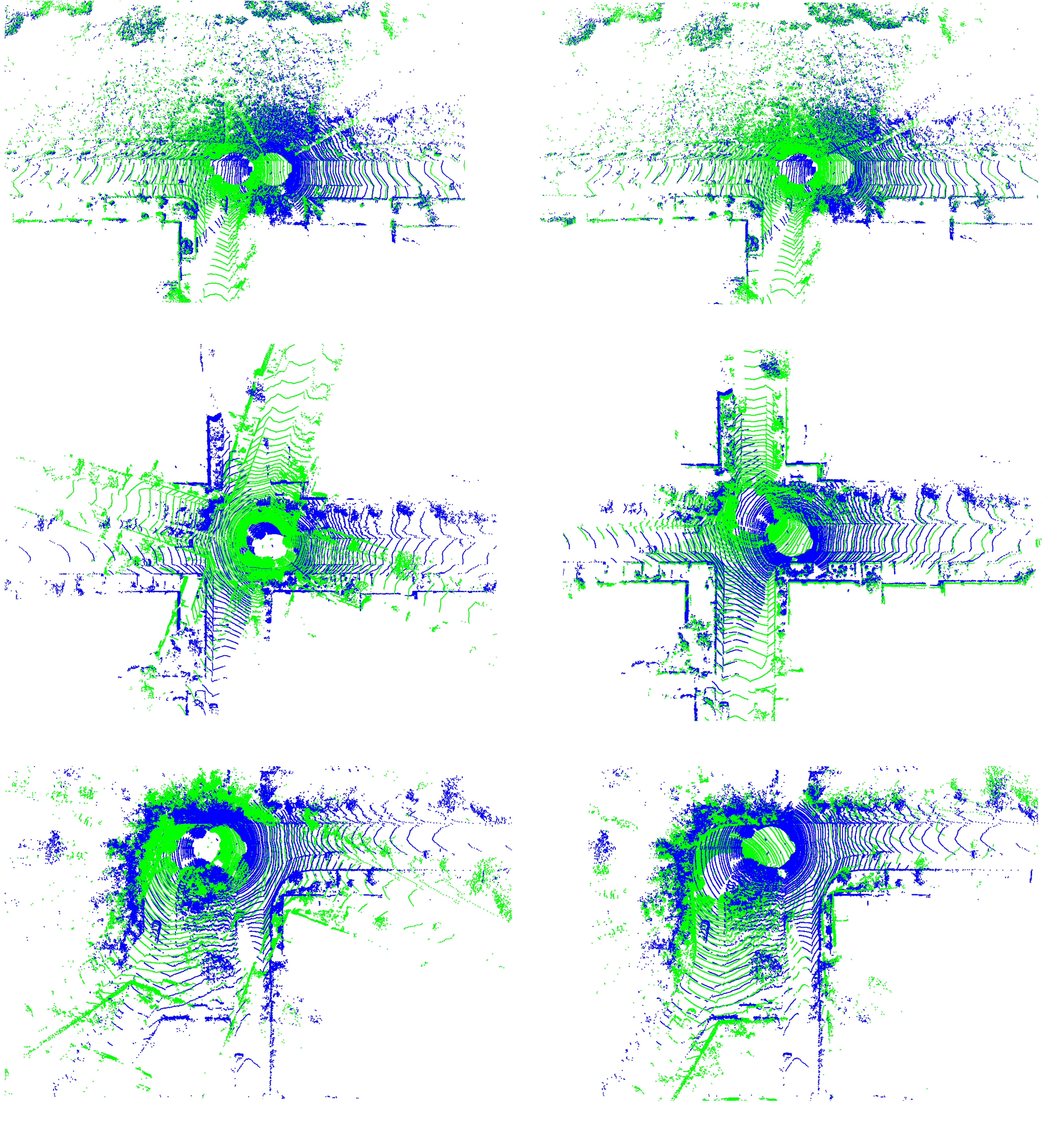}
\end{center}
\caption{Qualitative comparison of DGR~\cite{choy2020deep} (left column) and ours (right columne) on KITTI dataset. }
\label{fig:qualitative_kitti}
\end{figure*}

We attach additional qualitative comparisons with DGR~\cite{choy2020deep} on 3DMatch (\Fig{qualitative_3dmatch}) and on KITTI (\Fig{qualitative_kitti}).
It can be seen that our method performs well even for point cloud pairs which DGR fails to align accurately.
\clearpage
\clearpage

\end{document}